\documentclass[11pt,conference]{IEEEtran}
\IEEEoverridecommandlockouts
\usepackage{cite}
\usepackage{amsmath,amssymb,amsfonts}
\usepackage{algorithmic}
\usepackage{graphicx}
\usepackage{textcomp}
\usepackage{xcolor}

\usepackage{hyperref}
\usepackage[export]{adjustbox}
\usepackage[flushleft]{threeparttable} 
\graphicspath{{figs/}}
\def\etal{\emph{et al.}}

\title{Representing text as abstract images enables image classifiers to also simultaneously classify text}

\author{\IEEEauthorblockN{Stephen M.\ Petrie}
	\IEEEauthorblockA{\textit{Centre for Transformative Innovation} \\
		\textit{Swinburne University of Technology}\\
		Hawthorn, Australia \\
		spetrie@swin.edu.au}
	\and
	\IEEEauthorblockN{T'Mir D.\ Julius}
	\IEEEauthorblockA{\textit{Centre for Transformative Innovation} \\
		\textit{Swinburne University of Technology}\\
		Hawthorn, Australia \\
		tjulius@swin.edu.au}
}

\begin{document}
\maketitle
\begin{abstract}
	We introduce a novel method for converting text data into abstract image representations, which allows image-based processing techniques (e.g. image classification networks) to be applied to text-based comparison problems. We apply the technique to entity disambiguation of inventor names in US patents. The method involves converting text from each pairwise comparison between two inventor name records into a 2D RGB (stacked) image representation. We then train an image classification neural network to discriminate between such pairwise comparison images, and use the trained network to label each pair of records as either matched (same inventor) or non-matched (different inventors), obtaining highly accurate results. Our new text-to-image representation method could also be used more broadly for other NLP comparison problems, such as disambiguation of academic publications, or for problems that require simultaneous classification of both text and image datasets.
\end{abstract}

\section{Introduction}

Databases of patent applications and academic publications can be used to investigate the process of research and innovation. For example, patent data can be used to identify prolific inventors \cite{Gay2008} or to investigate whether mobility increases inventor productivity \cite{Hoisl2007}. However, the names of individuals in large databases are rarely distinct, hence individuals in such databases are not uniquely identifiable. For example, an individual named ``Chris Jean Smith'' may have patents under slightly different names such as ``Chris Jean Smith'', ``Chris J. Smith'', ``C J Smith'', etc\ldots\ There may also be different inventors with patents under the same or similar names, such as ``Chris Jean Smith'', ``Chris J. Smith'', ``Chris Smith'', etc\ldots\ Thus it is ambiguous which names (and hence patents) should be assigned to which individuals. Resolving this ambiguity and assigning unique identifiers to individuals --- a process often referred to as named entity disambiguation --- is important for research that relies on such databases.

Machine learning algorithms have been used increasingly in recent years to perform automated disambiguation of inventor names in large databases (e.g. \cite{Li2014,Ventura2015,Kim2016}). See Ventura \etal\ \cite{Ventura2015} for a review of supervised, semi-supervised, and unsupervised machine learning approaches to disambiguation. These more recent machine learning approaches have often out-performed more traditional rule- and threshold-based methods, but they have generally used feature vectors containing several pre-selected measures of string similarity as input for their machine learning algorithms. That is, the researcher generally pre-selects a number of string similarity measures which they believe may be useful as input for the machine learning algorithm to make discrimination decisions.

Here we introduce a novel approach of representing text-based data, which enables image classifiers to also simultaneously perform text classification. This new representation enables a supervised machine learning algorithm to learn its own features from the data, rather than selecting from a number of pre-defined string similarity measures chosen by the researcher. To do this, we treat the name disambiguation problem primarily as a classification problem --- i.e. we assess pairwise comparisons between records as either matched (same inventor) or non-matched (different inventors) \cite{Trajtenberg2006,Miguelez2011,Li2014,Ventura2015,Kim2016}. Then, for a given pairwise comparison between two inventor records, our text-to-image representation method converts the associated text strings into a stacked 2D colour image (or, equivalently, a 3D tensor) which represents the underlying text data.

\begin{figure*}
	\centering
	\includegraphics[width=0.15\textwidth, valign=c]{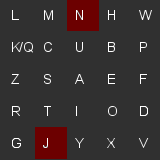}
	\space $+$ \space
	\includegraphics[width=0.15\textwidth, valign=c]{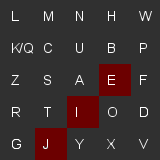}
	\space $+$ \space
	\includegraphics[width=0.15\textwidth, valign=c]{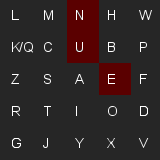}
	\space $+$ \space
	\includegraphics[width=0.15\textwidth, valign=c]{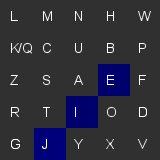}
	\space \space \space $=$ \space \space \space
	\includegraphics[width=0.15\textwidth, valign=c]{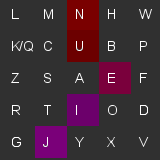}
	\caption{\textbf{Constructing a string-map image.} The first four images show the sub-maps that are summed to construct the final string-map image (right-most image), for the example word ``JEN''.}
	\label{fig:word-map}
\end{figure*}

We describe our text-to-image representation method in detail in Section \ref{sec:comparison-map_method} (see Figure \ref{fig:word-map} for an example of text-to-image conversion). We also test a number of alternative representations in Section \ref{sec:Testing different string maps}. Our novel method of representing text-based records as abstract images enables image processing algorithms (e.g. image classification networks), to be applied to text-based natural language processing (NLP) problems involving pairwise comparisons (e.g. named entity disambiguation). We demonstrate this by combining our text-to-image conversion method with a commonly used convolutional neural network (CNN) \cite{Krizhevsky2012}, obtaining highly accurate results (F1: 99.09\%, precision: 99.41\%, recall: 98.76\%).

\section{Related Work}

Inventor name disambiguation studies have often used measures of string similarity in order to make automated discrimination decisions. For example, counts of $n$-grams (sequences of $n$ words or characters) can be used to vectorise text, with the cosine distance between vectors providing a measure of string similarity \cite{Raffo2009,Pezzoni2014}. Measures of edit distance consider the number of changes required to transform one string to another, e.g.\ the number of additions, subtractions, or substitutions used in the calculation of Levenshtein distance \cite{Levenshtein1966}, or of other operations such as transpositions (the switching of 2 letters) used to calculate Jaro-Winkler distance \cite{Jaro1989,Winkler1990}. Phonetic algorithms, such as Soundex, recode strings according to pronunciation, providing a phonetic measure of string similarity \cite{Raffo2009}.

Measures of string similarity such as these have been used to guide rule- and threshold-based name disambiguation algorithms (e.g. \cite{Miguelez2011} and \cite{Morrison2017}). They can also be used within feature vectors inputted into machine learning algorithms. For example, Kim \etal\ \cite{Kim2016} use such string similarity feature vectors to train a random forest to perform pairwise classification. Ventura \etal\ \cite{Ventura2015} reviewed several supervised, semi-supervised, and unsupervised machine learning approaches to inventor name disambiguation, as well as implementing their own supervised approach utilising selected string similarity features as input to a random forest model.

Two-dimensional CNNs have been used extensively in recent image processing applications (e.g. \cite{Krizhevsky2012}), and one-dimensional (temporal) CNNs have been used recently as character-level CNNs for text classification (e.g. \cite{Zhang2015}). Also, neural networks (usually CNNs) have been used previously to assess pairwise comparison decisions --- e.g. in the case of pairs of: images \cite{Koch2015}, image patches \cite{Zbontar2016,Zagoruyko2015}, sentences \cite{Yin2016}, images of signatures \cite{Bromley1993}, and images of faces \cite{Hu2014}. These networks are generally constructed for multiple images to be provided simultaneously as input, such as in the case of siamese neural networks where two identical sub-networks are connected at their output \cite{Bromley1993,Koch2015}.

In this work we generate a \emph{single} 2-dimensional RGB (stacked) image for a given pairwise record comparison. Thus any image classification network that processes single images can be used (with minimal modification) to process our pairwise comparison images, therefore enabling such neural networks to also simultaneously classify associated text records. We demonstrate this using the seminal ``AlexNet'' image classification network \cite{Krizhevsky2012}.

\section{Data}

We use a combination of two labelled datasets in this work to train the neural network and assess its performance. Each dataset was derived by separate authors, from the US National Bureau of Economics Research (NBER) Patent Citation Data File \cite{Hall2001}; i.e. a labelled dataset of Israeli inventors \cite{Trajtenberg2006} (the ``IS'' dataset), and a dataset of patents filed by engineers and scientists \cite{Ge2016} (the ``E\&S'' dataset). These datasets were combined with US Patent and Trademark Office (USPTO) patent data as part of the PatentsView Inventor Disambiguation Workshop\footnote{http://www.patentsview.org/community/workshop-2015} hosted by the American Institutes for Research (AIR) in September 2015.

Each labelled dataset contains unique IDs that identify all inventor-name records from different patents belonging to each unique inventor. We also extracted several other variables from inventor-name records in the bulk USPTO patent data to use in our disambiguation algorithm: first name, middle name, last name, city listed in address, international patent classification (IPC) codes (i.e. subjects/fields covered by the patent), assignees (i.e. associated companies/institutes), and co-inventor names on the same patent.

\section{Disambiguation Algorithm}

Our novel inventor disambiguation algorithm involves the following main steps:

\begin{enumerate}
	\item \textbf{Duplicate removal:} remove duplicate inventor records.
	
	\item \textbf{Blocking:} block (or "bin") all names by last name, and also by first name in some cases.
	
	\item \textbf{Generate pairwise comparison-map images:} convert text from each within-block pairwise record comparison into a 2D RGB image representation.
	
	\item \textbf{Train neural network:} use 2D comparison-map images generated from manually labelled data to train a neural network to classify whether a given pairwise record comparison is a match (same inventor) or non-match (different inventors).
	
	\item \textbf{Classify pairwise comparison-map images:} deploy the trained neural network to classify pairwise comparison images generated from the bulk patent data, producing a match probability for each record pair.
	
	\item \textbf{Convert pairwise match probabilities into clusters:} convert the pairwise match/non-match probabilities generated by the neural net into inventor clusters --- i.e.\ groups of inventor-name records that each belong to a distinct individual inventor. Assigning a unique ID (UID) to each of these groups then leads to a single set of disambiguated inventor names.
\end{enumerate}

\noindent Note that the main purpose of the first two steps is to improve computational efficiency: i.e. rather than process all possible pairs of patent-inventor records (which has time complexity $\mathcal{O}(n^2)$ for $n$ records), the records are first grouped into similar clusters, or ``blocks'', and pairwise comparisons are only made within those blocks. For further detail regarding steps 1 and 2, see Appendices \ref{sec:deduplication} and \ref{sec:blocking}. Steps 3--6 are described in detail below.

\subsection{Comparison-map images}
\label{sec:comparison-map_method}

Our intent is to assess all possible within-block pairwise comparisons between patent-inventor records, classifying each comparison as either a match or non-match. To do this, we introduce a new method of converting any string of text into an abstract image representation of that text, which we refer to as a ``comparison-map'' image. Any image classification neural network can then be used to process these images and hence effectively perform text classification.

\begin{figure}
	\centering
	\includegraphics[width=0.12\textwidth, valign=c]{word-map.png}
	\space $+$ \space
	\includegraphics[width=0.12\textwidth, valign=c]{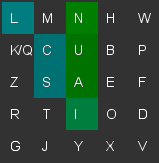}
	\space \space \space $=$ \space \space \space
	\includegraphics[width=0.12\textwidth, valign=c]{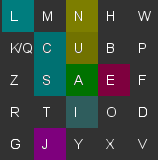}
	\caption{\textbf{Comparison of two strings.} To compare the names ``JEN'' and ``LIN'', we add the string-map for ``JEN'' (left image) to the string-map for ``LIN'' (middle image) to generate the final comparison image (right image).}
	\label{fig:word-map_JEN-LIN}
\end{figure}

To generate a comparison-map image, we firstly define a specific 2D character layout --- i.e. a grid of pixels specifying the positions of each letter. The layout of this ``string-map'' is shown in Figure \ref{fig:word-map} (identical in each of the five images). For a given word (e.g. ``JEN''), we then add a particular colour (e.g. red) to the pixels of each letter in the word, as well as to any pixels in straight lines connecting those letters. In particular, we add colour to the pixels of the first and last letters (Figure \ref{fig:word-map}, left-most image), and to all pixels in a line connecting each two-letter bi-gram (Figure \ref{fig:word-map}, second and third images, which correspond to the two bi-grams in ``JEN''; i.e. ``JE'' and ``EN''). To highlight the beginning of each string-map, we also repeat the process for the first bi-gram only (``JE'') in blue, rather than red (Figure \ref{fig:word-map}, fourth image). The final string-map for the word ``JEN'' is shown in Figure \ref{fig:word-map} (right-most image). If we then add the string-map of any other word to the green channel of the same RGB image (with the first bi-gram again highlighted in blue), the resulting image represents the pairwise comparison of the two words (e.g. Figure \ref{fig:word-map_JEN-LIN}, right-most image).

\begin{figure}
	\centering
	\includegraphics[width=0.22\textwidth]{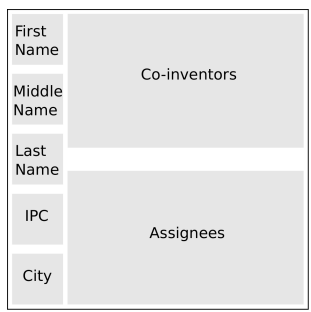}
	\caption{\textbf{Record-map layout.} Shows the positioning of each string-map within a given record-map.}
	\label{fig:record-map}
\end{figure}

For a given inventor name record, we generate string-maps for each variable in the record --- i.e. first name, middle name, last name, city, IPC codes, co-inventors, and assignees. These string-maps are combined into a single image, arranged as shown in Figure \ref{fig:record-map}, which we refer to as a ``record-map''.

\begin{figure}
	\centering
	\includegraphics[width=0.37\textwidth]{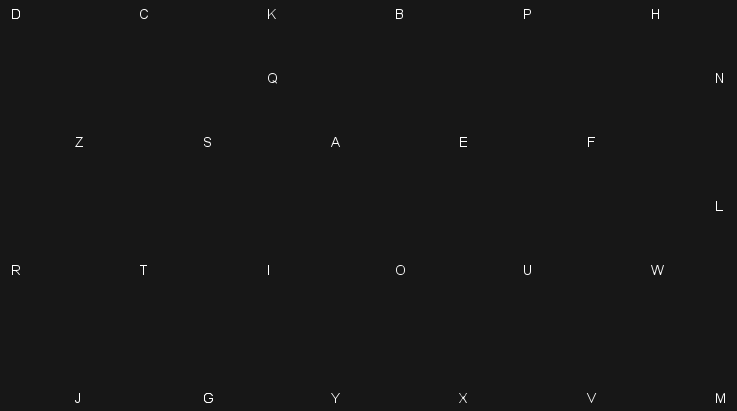}
	\includegraphics[width=0.1\textwidth]{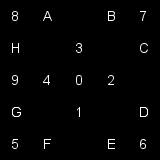}
	\caption{\textbf{Larger string-map for assignees and co-inventors, and IPC-map.} The larger string-map used to convert a given list of assignees or co-inventors into an abstract image representation (left), and the IPC-map used to convert a given list of IPC classes into an abstract image representation (right).}
	\label{fig:word-map_stretched_ipc-map}
\end{figure}

Since a given patent-inventor record can have multiple assignees and/or co-inventors, we use a larger string-map for those variables (see Figure \ref{fig:word-map_stretched_ipc-map}, left image). This reduces the possibility that pixels will become saturated in cases where many assignees (or co-inventors) are overlayed onto the same string-map. We also add less colour to each pixel in these larger string-maps, again to reduce the possibility of saturation. For international patent classification (IPC) codes, which contain numbers as well as letters, we use a different string-map shown in Figure \ref{fig:word-map_stretched_ipc-map} (right image).

\begin{figure*}
	\begin{center}
		\includegraphics[width=0.49\textwidth]{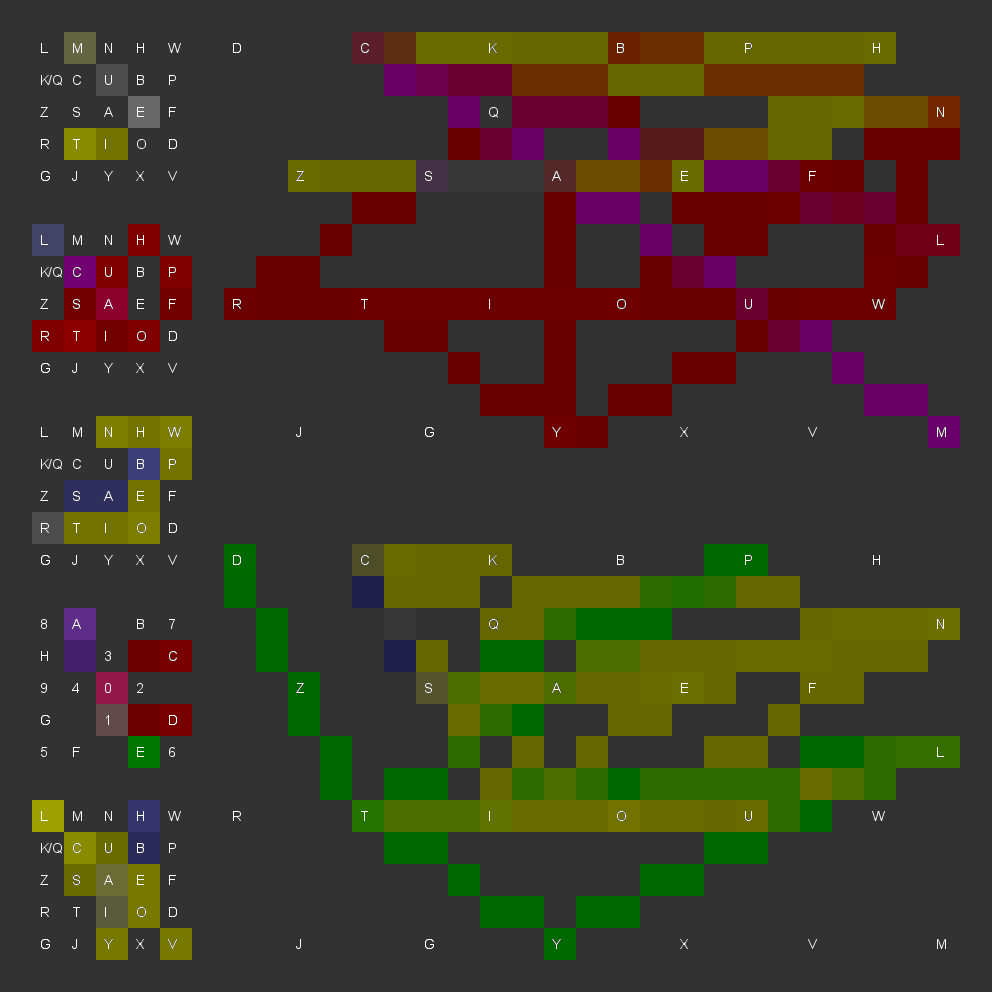}
		\includegraphics[width=0.49\textwidth]{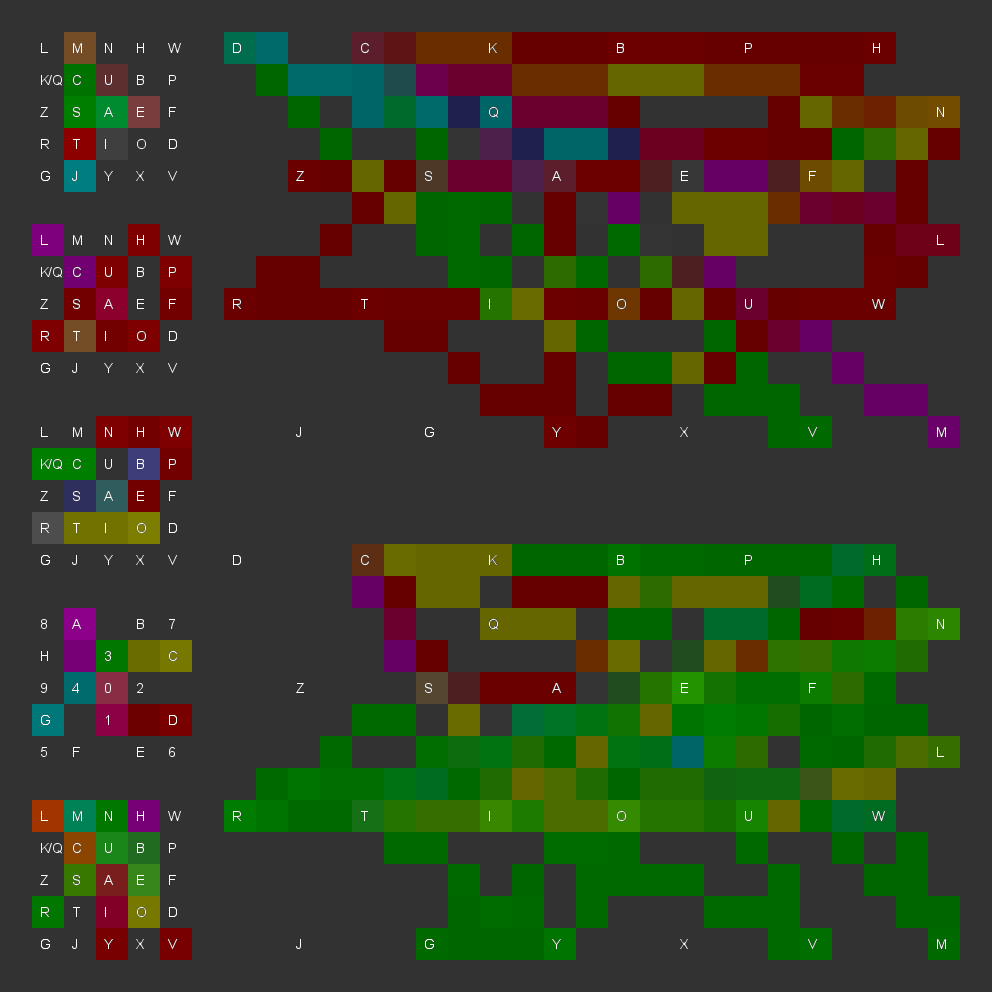}
	\end{center}
	\caption{\textbf{Comparison-map examples.} Two examples of comparison-map images. The left comparison-map image was generated using two matched records (Table \ref{tab:mock_records}, rows 1 and 2), and the right image from two non-matched records (Table \ref{tab:mock_records}, rows 1 and 3).}
	\label{fig:comparison-map}
\end{figure*}

We compare any two inventor name records by stacking the two associated 2D record-maps into the same RGB image, one as the red channel and the other as green (with the beginning two-letter bi-gram of each record sharing the blue channel). We refer to the resulting RGB image (or 3D tensor) representation as a ``comparison-map'' (Figure \ref{fig:comparison-map}).

Since red and green combined produce yellow in the RGB colour model, a comparison-map image generated from two similar records should contain more yellow (e.g. Figure \ref{fig:comparison-map}, left image), whereas a comparison-map image from two dissimilar records should contain more red and green (e.g. Figure \ref{fig:comparison-map}, right image) due to less overlap between the two record-maps. When training on labelled comparison-maps, we expect that the neural network will learn to identify features such as these, which are useful for discriminating between matched/non-matched records. That is, the neural network's learned pattern recognition on comparison-map images will essentially recognise underlying text patterns which are present in the associated patent-inventor name records.

Note that we chose the particular layout of the letters in the string-map shown in Figure \ref{fig:word-map} heuristically, such that vowels (which are less important than consonants when assessing string similarity) are positioned towards the centre of the grid, where pixels are more likely to saturate. We also grouped letters with similar phonetic interpretations, such as ``S'' and ``Z'', close to each other. We anticipated that this heuristic layout might make it more straightforward for the network to learn which features are associated with matches/non-matches. However, we test how the heuristic layouts shown in Figures \ref{fig:word-map}---\ref{fig:word-map_stretched_ipc-map} perform compared with alternative random layouts later in Section \ref{sec:Testing different string maps}, and find similar performance regardless of the chosen layout.

Our method of converting text into a stacked 2D RGB bitmap for neural net-based image classification has several benefits:

\begin{itemize}
	\item The powerful classification capabilities of previous image classification networks can be utilised for text-based record matching, with minimal modification.
	\item The neural network learns its own features from the data, rather than learning from a feature vector of pre-defined string similarity measures chosen by the researcher.
	\item Minor spelling variations and errors do not alter the resulting string-map very much, and the neural network can potentially learn that such minor features are unimportant for discriminating between matches and non-matches.
	\item Matched records with differing word ordering (e.g. re-ordered co-inventor names on different patents) are likely to be matched, due to overlapping pixels.
	\item The neural net can potentially learn to ignore certain shapes of common words (e.g. ``Ltd'', ``LLC'', ``Inc'', etc\ldots) which are not useful for discrimination decisions.
	\item Our novel disambiguation algorithm performs well under multiple different choices of alternative string-maps other than those shown in Figures \ref{fig:word-map}---\ref{fig:word-map_stretched_ipc-map} (see Section \ref{sec:Testing different string maps}), suggesting that multiple alternatives of our comparison-map representations allow for robust pattern recognition and feature extraction.
\end{itemize}

\noindent Note that the above benefits of our text-to-image conversion method would also apply to other text-based comparison problems (e.g. data linkage, or disambiguation of academic papers), or to problems that require simultaneous classification of both text and image datasets.

\begin{table*}
	\small
	\caption{Mock records of three patent-inventor name instances. Rows 1 and 2 are the same mock inventor, while row 3 is a different inventor.}
	\label{tab:mock_records}
	\begin{tabular}{llll p{4.7cm}}
		\hline\noalign{\smallskip}
		Name & IPC codes & City & Co-inventors (last names) & Assignees \\
		\hline\noalign{\smallskip}
		Emmett Lathrop Brown & A10C, A10D & Hill Valley & McFly, Clayton-Brown, Sanchez & Science Solutions \\
		Emmett L. Brown & A11E & Hill Valley & Sanchez & Science Solutions Pty. Ltd. \\
		James T. Brock & G03C & Melbourne & Edison, Da Vinci & Swinburne University of Technology, The University of Melbourne \\
		\hline
	\end{tabular}
\end{table*}

\subsection{Modifications to neural network architecture}

\begin{table*}
	\small
	\begin{center}
		\begin{tabular}{p{5cm} c c p{4cm}}
			\hline\noalign{\smallskip}
			Hyperparameter & AlexNet & This work & Rationale for modification \\
			\hline\noalign{\smallskip}
			Number of neurons in input layer & $224 \times 224 \times 3 = 150,528$ & $31 \times 31 \times 3 = 2,883$ & Smaller size of input images \\
			Kernel size in first convolutional layer & $11 \times 11 \times 3$ & $3 \times 3 \times 3$ & Smaller-scale features to learn \\
			Stride length of kernels in 1st conv layer & 4 & 1 & Smaller kernel size \\
			Number of neurons in output layer & 1,000 & 2 & Fewer classes \\
			\hline
		\end{tabular}
	\end{center}
	\caption{Hyperparameters that differ between the two network architectures. See \cite{Krizhevsky2012} for more details on the network architecture.}
	\label{tab:network_hyperparams}
\end{table*}

To demonstrate that our text-to-image conversion method can be combined with an image classifier to perform text-based classification, we apply the method to a commonly used image classification neural network; i.e. the seminal ``AlexNet'' CNN \cite{Krizhevsky2012}. AlexNet was originally designed to classify colour images (224$\times$224$\times$3-pixel bitmaps) amongst 1,000 classes. We slightly modify the network architecture to enable classification of pairwise comparison-map images (31$\times$31$\times$3-pixel bitmaps) into two classes (match/non-match), by altering four hyperparameters as shown in Table \ref{tab:network_hyperparams}. We use the NVIDIA Deep Learning GPU Training System\footnote{\href{https://developer.nvidia.com/digits}{https://developer.nvidia.com/digits}} (DIGITS) v2.0.0 implementation of AlexNet, and use the Caffe backend \cite{Jia2014}. We use the default settings for the DIGITS solver (stochastic gradient descent), batch size (100), and number of training epochs (30). Rather than use the default learning rate (0.01), we use a sigmoid decay function to progressively decrease the learning rate from 0.01 to 0.001 over the course of the 30 training epochs, as testing indicated that this produced slightly higher accuracies. Instead of the 1,000-neuron softmax output layer in AlexNet, we use a 2-neuron softmax output layer, which outputs a probability distribution across our two possible classes (match/non-match).

Note that the default settings of the DIGITS v2.0.0 implementation of AlexNet transform the input data by: (1) altering input images to show the deviation from the mean of all input images (by subtracting the mean image from each input image); (2) randomly mirroring input images; and (3) taking a random square crop from the input image. The main purpose of performing such transformations is to introduce variability into the training images that are expected to be present in the unlabelled data, however we do not use any of those transformations in this work because our images are much more self-consistent than those in the ImageNet database.

\subsection{Converting pairwise probabilities into inventor groups, and assigning UIDs}

After running the trained neural network on bulk patent data, each within-block pairwise comparison has an associated match probability. To assign unique IDs (UIDs) to the bulk data, we convert these pairwise probabilities into linked (matched) ``inventor groups'' using a clustering algorithm. Each inventor group is a linked cluster of inventor name records which all refer to the same individual. Briefly, the clustering algorithm involves converting each pairwise probability value to a binary value (match/non-match) using a pre-selected probability threshold ($\bar{p}$) as a cut-off. Each matched record is then clustered into a larger inventor group if the number of links ($l$) it has to the that group is $\geqslant$ the number of nodes in the group ($n$) times some threshold proportion value ($\bar{l}$); i.e. if $l \geqslant n \bar{l}$. This removes weakly-linked records from each group. For further detail on the clustering algorithm, see Appendix \ref{sec:clustering}. Note that choosing different $\bar{p}$ and $\bar{l}$ values generates different trade-offs between precision and recall. Once the clustering algorithm has been applied to each block, every patent-inventor name instance has an associated unique inventor ID, and the disambiguation process is complete.

\section{Results}

Here we firstly describe our procedure for dividing our labelled datasets into training and test data. We then evaluate our inventor disambiguation algorithm, compare those results to previous studies, and test alternative string-map layouts.

\subsection{Labelled and bulk datasets}

We use the IS and E\&S labelled datasets to train the neural network to discriminate between matched and non-matched pairwise comparisons. Each of the labelled datasets are randomly separated into 80\% training data (used to train the neural network) and 20\% test data (used to assess algorithm performance). We use 75\% of the training data to train the network, and the remaining 25\% to perform validation assessments during training in order to monitor potential overfitting.

Duplicate removal and blocking is then performed on the labelled data, and comparison-map images are generated for all possible pairwise record comparisons within each block (723,178 comparison-maps for training and 144,552 comparison-maps for testing).

We also perform duplicate removal and blocking on the bulk data, generating comparison-maps for all possible pairwise within-block comparisons (stored as 3D numerical arrays). The trained neural network is then deployed on the bulk patent data, generating match/non-match probabilities for all pairwise within-block comparisons (112,068,838 comparison-maps). Prior to processing the bulk data, we experimented with multiple different values for the pairwise comparison probability threshold ($\bar{p}$) and linking proportion threshold ($\bar{l}$), based on evaluating the trained neural network on the labelled test data. Different $\bar{p}$ and $\bar{l}$ values produce different trade-offs between precision and recall, and we use values that produce an optimal trade-off (highest F1 score). We state each $\bar{p}$ and $\bar{l}$ value whenever quoting results from a given run of our disambiguation algorithm.

\subsection{Evaluation}

To evaluate the performance of the disambiguation algorithm, we use the labelled IS and E\&S test data to estimate pairwise precision, recall, splitting, and lumping based on numbers of true positive (\emph{tp}), false positive (\emph{fp}), true negative (\emph{tn}), and false negative (\emph{fn}) pairwise links within the labelled test data, as follows (e.g. \cite{Ventura2015,Kim2016}):

\begin{equation}
\emph{Precision} = {{\emph{true pos.\ matches}} \over {\emph{all pos.\ matches}}} = {{\emph{tp}} \over {\emph{tp} + \emph{fp}}}
\end{equation}

\begin{equation}
\emph{Recall} = {{\emph{true pos.\ matches}} \over {\emph{total true matches}}} = {{\emph{tp}} \over {\emph{tp} + \emph{fn}}}
\end{equation}

\begin{equation}
\emph{Splitting} = {{\emph{false neg.\ non-matches}} \over {\emph{total true matches}}} = {{\emph{fn}} \over {\emph{tp} + \emph{fn}}}
\end{equation}

\begin{equation}
\emph{Lumping} = {{\emph{false pos.\ matches}} \over {\emph{total true matches}}} = {{\emph{fp}} \over {\emph{tp} + \emph{fn}}}
\end{equation}

\noindent Higher values are better for precision and recall, while lower values are better for lumping and splitting errors. We also use the pairwise F1 score:

\begin{equation}
\emph{F1} = 2 \times {{\emph{Precision} \cdot \emph{Recall}} \over {\emph{Precision} + \emph{Recall}}}
\end{equation}

\noindent Since the F1 score accounts for the trade-off between precision and recall, it is the primary measure we use to compare the performance of different disambiguation algorithms.

\subsection{Disambiguation algorithm performance}
\label{sec:Disambiguation algorithm performance}

\begin{table}
	\centering
	\small
	\caption{Performance of two example runs of our disambiguation algorithm (bottom rows), compared with other studies evaluated on the IS or E\&S labelled datasets. All values in \%.}
	\label{tab:results_is-ens-only}
	\begin{threeparttable}
		\begin{tabular}{p{3cm} cccc}
			\hline\noalign{\smallskip}
			Method & [$\bar{p}$; $\bar{l}$]	&   Recall  &   Precision   &   F1	\\
			\noalign{\smallskip}\hline\noalign{\smallskip}
			Kim2016 \cite{Kim2016}; IS   & &   98.13  &  99.89  &   99.00 \\
			Kim2016 \cite{Kim2016}; E\&S   & &  98.10   &  99.95  &   99.02 \\
			Kim2016 \cite{Kim2016}; Both\tnote{$\dagger$}   & &  98.12   &  99.92  &   99.01 \\
			\noalign{\smallskip}
			Yang2017 \cite{Yang2017}; IS   & &   83.79  &  99.57  &   91.00 \\
			Yang2017 \cite{Yang2017}; E\&S   & &  90.31   &  99.87  &   94.85 \\
			Yang2017 \cite{Yang2017}; Both\tnote{$\dagger$}   & &  87.05   &  99.72  &   92.93 \\
			\noalign{\smallskip}
			Ours; Both   & [0.02; 0.1]   &  98.67 &   99.48 &   99.07 \\
			Ours\tnote{*}\ ; Both    & [0.03; 0.05]   &   98.76  &   99.41 &   \textbf{99.09} \\
			\noalign{\smallskip}\hline
		\end{tabular}
		\begin{tablenotes}
			\item[$\dagger$] Calculated by averaging the IS and E\&S results.
			\item[*] Note that this result was obtained using a randomly-generated string-map character order (see Section \ref{sec:Testing different string maps}).
		\end{tablenotes}
	\end{threeparttable}
\end{table}

\begin{table*}
	\centering
	\small
	\caption{Performance of our disambiguation algorithm relative to other studies, regardless of evaluation dataset. All values in \%.}
	\label{tab:results}
	\begin{threeparttable}
		\begin{tabular}{lcccccc}
			\hline\noalign{\smallskip}
			Method	& [$\bar{p}$; $\bar{l}$]	&   Splitting	&	Lumping    &   Recall  &   Precision   &   F1	\\
			\noalign{\smallskip}\hline\noalign{\smallskip}
			Li2014 \cite{Li2014}\tnote{$\dagger$}   & &   3.26    &   2.34 &      &      &    \\
			\noalign{\smallskip}
			Ventura2015 \cite{Ventura2015}   & &   2.31    &   1.64    &     &     &    \\
			\noalign{\smallskip}
			Kim2016 \cite{Kim2016}  & &   &    &   98.48  &   99.60 &   99.04 \\
			\noalign{\smallskip}
			Morrison2017 \cite{Morrison2017}  & &   &    &   92  &   98 &   95 \\
			\noalign{\smallskip}
			Yang2017 \cite{Yang2017}  & &   &    &   96.15  &   99.61 &   97.85 \\
			\noalign{\smallskip}
			Ours	& [0.03; 0.05]   &  1.24  &  0.58  &   98.76  &   99.41 &   \textbf{99.09} \\
			\noalign{\smallskip}\hline
		\end{tabular}
		\begin{tablenotes}
			\item[$\dagger$] Ventura \etal\ \cite{Ventura2015} also use an ``optoelectronics'' (OE) labelled dataset to evaluate \cite{Li2014}, obtaining lower errors on the full OE dataset (splitting: 2.49\%, lumping: 0.39\%), but higher errors on a random sample of OE data (splitting: 10.54\%, lumping: 1.21\%).
		\end{tablenotes}
	\end{threeparttable}
\end{table*}

\begin{table*}
	\centering
	\small
	\caption{Comparison of alternate string-map layouts. Each row shows the highest F1 result obtained for that string-map layout.}
	\label{tab:alternative_stringmap_results}
	\begin{tabular}{llccc}
		\hline\noalign{\smallskip}
		String-map layout	& [$\bar{p}$; $\bar{l}$]	&   Recall  &   Precision   &   F1 \\
		\noalign{\smallskip}\hline\noalign{\smallskip}
		Heuristic character order \& layout	& [0.02; 0.1]  &   98.67 &   99.48 &   99.07 \\
		Random order, heuristic layout	& [0.03; 0.05] &   98.76  &   99.41  &   \textbf{99.09} \\
		Random order \& layout	& [0.05; 0.05]	&   98.77  &   99.29 &   99.03 \\
		Random order \& layout, with small string-maps	& [0.05; 0.2]   &  98.46  &   99.52 &  98.99 \\
		Random order, heuristic layout, no blue channel	& [0.02; 0.05]   &  98.71  &   99.32 &  99.01 \\
		\noalign{\smallskip}\hline
	\end{tabular}
\end{table*}

The precision, recall, and F1 estimates for two example runs of our disambiguation algorithm are shown in the bottom two rows of Table \ref{tab:results_is-ens-only} --- first is the highest F1 result obtained using the heuristic string-map character order (Figures \ref{fig:word-map}---\ref{fig:word-map_stretched_ipc-map}), and second is the highest F1 result obtained using a randomly-generated string-map character order (see Section \ref{sec:Testing different string maps} for details). Table \ref{tab:results_is-ens-only} also shows the best results (highest F1) obtained by previous studies which (1) disambiguate bulk USPTO patent data, and (2) evaluate their results using the same labelled datasets we use in this work (i.e. the IS and E\&S datasets). Our inventor disambiguation algorithm performs well compared with these other disambiguation studies (Table \ref{tab:results_is-ens-only}, bottom row), marginally out-performing the previous state-of-the-art study of Kim \etal\ \cite{Kim2016} and obtaining a much higher F1 score than Yang \etal\ \cite{Yang2017}.

For completeness, we also compare our results to those of other studies which use alternative labelled datasets to the IS and E\&S datasets used in this work --- i.e. Table \ref{tab:results} shows the best results obtained by each study, regardless of the evaluation dataset. Note that Table \ref{tab:results} provides a less equitable comparison than Table \ref{tab:results_is-ens-only}, as there is generally a small amount of variation in an algorithm's F1 score when evaluated on different labelled datasets. Nonetheless, we include Table \ref{tab:results} here for completeness and consistency with previous inventor name disambiguation studies, which often include comparison to other studies with different evaluation datasets. Our disambiguation algorithm is again competitive with the other state-of-the-art inventor name disambiguation algorithms in Table \ref{tab:results}, obtaining the highest F1 score compared with the other three studies which quote F1 results, and lower splitting and lumping errors compared with the \cite{Li2014} and \cite{Ventura2015} studies (those studies did not quote F1 results).

\subsection{Testing alternative string-maps}
\label{sec:Testing different string maps}

Here we compare the performance of our heuristic string-map layouts (Figures \ref{fig:word-map}---\ref{fig:word-map_stretched_ipc-map}) to several alternative string-maps. The first alternative string-map we test has random character order; i.e. we keep the pixel co-ordinates identical to the co-ordinates of the associated heuristic layout, but randomise the order of each character (these randomised string-maps are shown in Appendix \ref{sec:Random string-map layouts}, Figure \ref*{fig:maps_random-charpositions}). We also test two alternative string-maps in which we randomise both the pixel co-ordinate layout and character order (Appendix \ref{sec:Random string-map layouts}, Figure \ref*{fig:maps_random-charpositions-and-layout}). One alternative uses the large string-map for co-inventors and assignees (Figure \ref*{fig:maps_random-charpositions-and-layout}, bottom image). The other alternative uses the smaller $5 \times 5$ pixel string-map for co-inventors and assignees (Figure \ref*{fig:maps_random-charpositions-and-layout}, top-left image), leading to a smaller comparison-map (Appendix \ref{sec:Random string-map layouts}, Figure \ref*{fig:compmap_random-charpositions-and-layout_19x19}). We also investigate a string-map with random character order in which we exclude the blue channel for leading bi-grams (Figure \ref{fig:word-map}, fourth image).

Estimates of precision, recall, and F1 for each of these alternative string-maps are shown in Table \ref{tab:alternative_stringmap_results}. For each alternative string-map, we ran the algorithm multiple times using different settings of the comparison probability threshold ($\bar{p}$) and linking proportion threshold ($\bar{l}$), and only show results from the run which produced the highest F1 score. Results obtained from each of the alternative string-maps are quite similar to those obtained using the heuristically-determined layout (F1 scores range from 98.99\% to 99.09\%). This suggests that our method of converting text into abstract image representations facilitates robust feature learning for several alternative choices of string-map structure, such as randomised string-map character order and/or layout, heuristic order and/or layout, different string-map sizes, and the inclusion/exclusion of a blue channel for leading bi-grams.

\section{Conclusion}

Our name disambiguation algorithm provides a novel way of combining image processing with NLP, allowing image classifiers to perform text classification. We demonstrated this with the AlexNet CNN, producing highly accurate results. We also analysed several variants of alternative string-maps, and found that the accuracy of the disambiguation algorithm was highly robust to such variation.

Our disambiguation algorithm could easily be adapted to other NLP problems requiring text matching of multiple strings, such as academic author name disambiguation or record linkage problems. The algorithm could also potentially be modified to process records that contain both text and image data, by combining each record's associated image with the abstract image representation of the record's text, in a single comparison-map.

\section*{Acknowledgment}

We thank Assoc.\ Prof.\ Kai Qin and Prof.\ Elizabeth Webster for their helpful comments on the manuscript.

\bibliography{name_matching}
\bibliographystyle{IEEEtran}

\setcounter{figure}{0}
\renewcommand{\thefigure}{A\arabic{figure}}

\appendix

\subsection{Removal of duplicate records}
\label{sec:deduplication}

It is sometimes obvious that two inventor name records likely belong to the same individual, because the two records contain several fields that are identical. For example, if the last name, first name, city, and IPCs of two different records are all exactly identical, it is highly likely that the two records belong to the same individual. We remove such duplicate records based on the following duplication keys:

\begingroup
\begin{verbatim}
duplicnkey_ipc = lastname + firstname
+ city + '_'.join(ipcs)
\end{verbatim}
\begin{verbatim}
duplicnkey_assignee = lastname + firstname
+ city + '|'.join(assignees)
\end{verbatim}
\endgroup

\noindent For a given group of duplicate records sharing the same duplication key, all records except for the first record to be processed are removed from the bulk data. The first record then remains within the bulk data to be processed by the disambiguation algorithm, receiving a unique inventor ID once the algorithm has completed its run. That same ID is then assigned to each removed record in the corresponding group of duplicate records.

\subsection{Blocking}
\label{sec:blocking}

The blocking  procedure broadly involves grouping together inventor name records into ``blocks'' (or ``bins'') using each inventor's last name, and sometimes also their first name. Latter parts of the algorithm will only assess pairwise comparisons within these blocks, never across different blocks.

We firstly group patent-inventor name records together by the first three letters of the last name (this first step is identical to the initial stage of the blocking procedure used by Ventura \emph{et al.}\ \cite{Ventura2015}). However, some of the resulting blocks contain very large numbers of records, and hence large numbers of pairwise comparisons. To improve efficiency, we further divide such large blocks into smaller blocks by progressively increasing the number of letters used for blocking. That is, if the number of records within a given block ($n_b$) is above some threshold number ($\bar{n}_b$), then the records within that block are separated into smaller blocks according to the first \emph{four} letters of the last name. We then continue sub-dividing any blocks that still have $n_b > \bar{n}_b$, according to the first five letters of the last name, then six letters, and so on. If all letters of the last name have been used and any blocks still have $n_b > \bar{n}_b$, then we append a comma to the string and begin progressively appending letters from the first name as well.

We use $\bar{n}_b = 100$ throughout this work, as initial testing indicated that it produced a good balance between:

\begin{itemize}
	\item computational efficiency: i.e. smaller $\bar{n}_b$ leads to more numerous, smaller bins; hence fewer comparisons (which are $\mathcal{O}(n_b^2)$ for each bin) and less computation time,
	\item accuracy: i.e. smaller $\bar{n}_b$ reduces the number of unnecessary comparisons between records (often non-matched records) which should reduce false positives,
	\item recall: i.e. larger $\bar{n}_b$ leads to fewer, larger bins, which decreases splitting errors (decreasing false negatives),
\end{itemize}

\noindent Together with the deduplication procedure, this reduces the number of pairwise comparisons from $\approx 77$ trillion before the blocking procedure to $\approx 112$ million.

Note that since latter parts of the algorithm only assess within-block pairwise comparisons and some inventors' sets of records may have been separated across two or more different blocks, there is a maximum limit to the possible recall attainable by the disambiguation algorithm. After running the blocking procedure on the labelled dataset, we use known pairwise matches in the labelled data to estimate this maximum limit to recall, obtaining the following values: 99.47\% (E\&S training data), 99.98\% (E\&S test data), 99.83 (IS training data), and 99.86 (IS test data).

\subsection{Clustering algorithm to assign inventor groups}
\label{sec:clustering}

Here we describe the clustering algorithm we use to convert pairwise match probabilities into groups of records each belonging to a single unique inventor. We firstly convert each pairwise probability between the $i$th and $j$th record ($p_{ij}$) into one of the binary classes ($c_{ij}$; either ``match'' or ``non-match'') based on a threshold probability value ($\bar{p}$) as follows:

\begin{equation}
c_{ij} =
\begin{cases}
\text{match} \: ,	& \text{if $p_{ij} \geqslant \bar{p}$} \\
\text{non-match} \: ,	& \text{otherwise.}
\end{cases}
\end{equation}

The inventor group linking algorithm then primarily involves combining different sub-groups together into the one group if they share enough links (pairwise matches). Within a given block, the algorithm involves the following steps:

\begin{enumerate}
	\item Order all patent-inventor name records by the number of links they have to other records (i.e. the number of asserted matches to other records), highest first.
	
	\item Assign a UID to each isolated (non-matched) patent-inventor name.
	
	\item Assign records to inventor groups. That is, for a given record, the corresponding inventor group initially comprises just the record itself and all records it is linked (matched) to. Each of these linked records (nodes) are kept in the current inventor group only if the number of links ($l$) it has to the current group is $\geqslant$ the number of nodes in the group ($n$) times some threshold proportion ($\bar{l}$); i.e. if $l \geqslant n \bar{l}$. This removes the most weakly-linked records from each group (i.e. the nodes with fewest links to their group), which are more likely to be false positive matches. Any outside-group links --- i.e. links to nodes that are not within the current group --- are also recorded during this step.
	
	\item Repeat Step 2, because some records may have become isolated (non-matched) following Step 3.
	
	\item Combine inventor groups together if the number of links they share is greater than a specified threshold. In particular, for an inventor group with $n_{\text{self}}$ records (nodes), we combine it with any other group with $n_{\text{other}}$ nodes if the number of links to that other group ($l$) satisfies both: $l \geqslant \bar{l} \, n_{\text{self}}$, and: $l \geqslant \bar{l} \, n_{\text{other}}$.
	
	\item For each resulting inventor group, assign an identical UID to all patent-inventor name records in the group.
\end{enumerate}

\subsection{Random string-map layouts}
\label{sec:Random string-map layouts}

\begin{figure}
	\centering
	\includegraphics[width=0.09\textwidth]{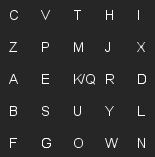}
	\includegraphics[width=0.09\textwidth]{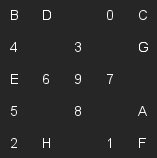}
	\includegraphics[width=0.37\textwidth]{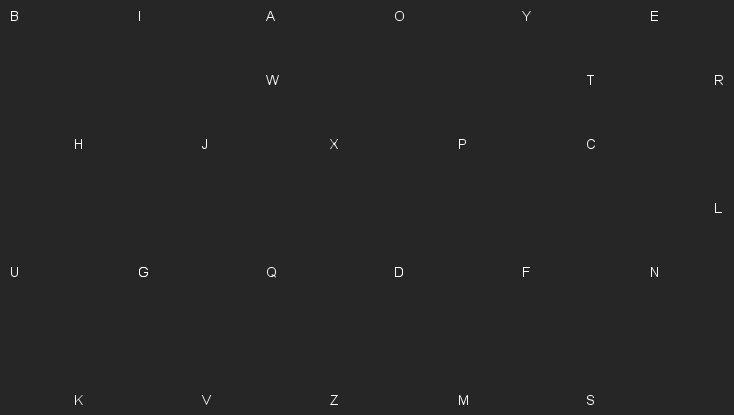}
	\caption{\textbf{Random character order.} Here we show the smaller string-map (top-left), IPC-map (top-right), and larger string-map (bottom) we use for runs in which the character order has been randomised.}
	\label{fig:maps_random-charpositions}
\end{figure}

Here we show the random string layouts analysed in Section \ref{sec:Testing different string maps}. Figure \ref*{fig:maps_random-charpositions} shows the string-maps we use for runs where characters are positioned using an identical pixel co-ordinate layout to the heuristic layouts shown in Figures \ref{fig:word-map} and \ref{fig:word-map_stretched_ipc-map} (main text), but where the order of each character has been randomised.

\begin{figure}
	\centering
	\includegraphics[width=0.09\textwidth]{stringmap_random-charpositions.png}
	\includegraphics[width=0.09\textwidth]{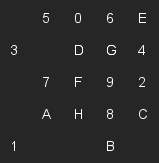}
	\includegraphics[width=0.37\textwidth]{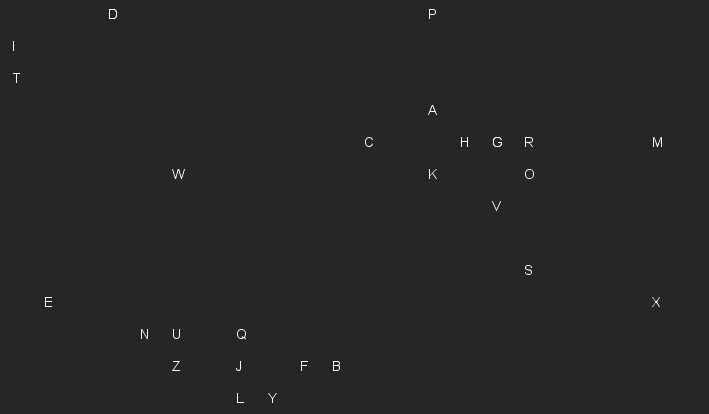}
	\caption{\textbf{Random character order and layout.} Here we show the smaller string-map (top-left; identical to the top-left string-map in Figure \ref*{fig:maps_random-charpositions}), IPC-map (top-right), and larger string-map (bottom) with both random character order and random pixel co-ordinate layout.}
	\label{fig:maps_random-charpositions-and-layout}
\end{figure}

Figure \ref*{fig:maps_random-charpositions-and-layout} shows the string-maps used for runs where both pixel co-ordinate layout and character order are randomised.

\begin{figure}
	\centering
	\includegraphics[width=0.34\textwidth]{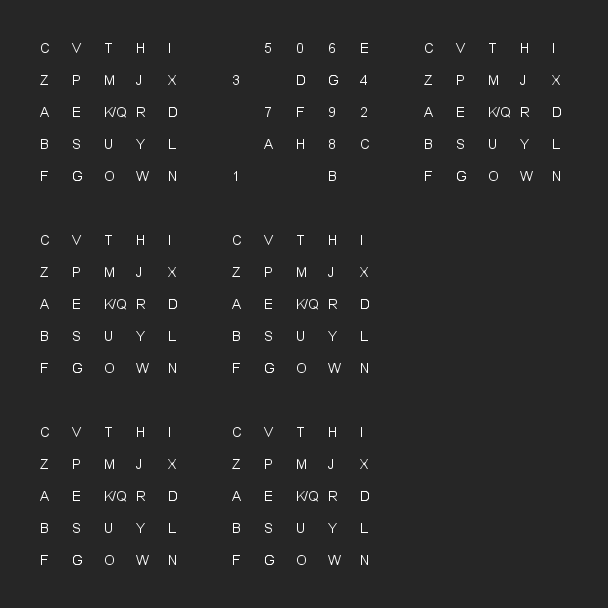}
	\includegraphics[width=0.14\textwidth]{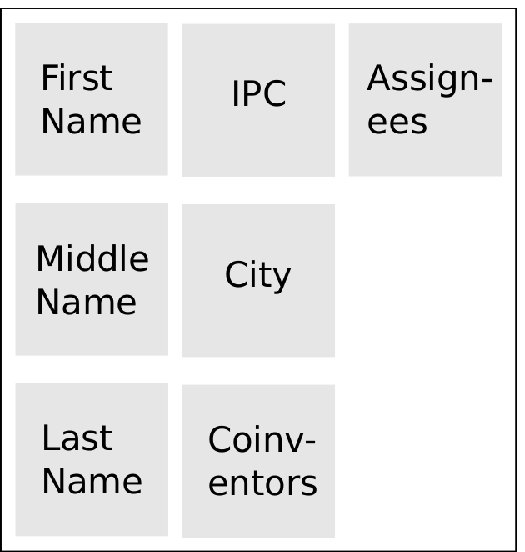}
	\caption{\textbf{Random character order and layout, with small string-maps.} The left image shows the comparison-map used for runs with smaller string-maps for co-inventors and assignees, as well as random character order and random pixel co-ordinate layout. The right image shows the associated record-map layout.}
	\label{fig:compmap_random-charpositions-and-layout_19x19}
\end{figure}

The left image in Figure \ref*{fig:compmap_random-charpositions-and-layout_19x19} shows the comparison-map with random layout and character order in which we use the smaller $5 \times 5$ pixel string-map (Figure \ref*{fig:maps_random-charpositions-and-layout}, top-left image) for co-inventors and assignees, rather than the larger string-map (Figure \ref*{fig:maps_random-charpositions-and-layout}, bottom image). The right image in Figure \ref*{fig:compmap_random-charpositions-and-layout_19x19} shows the record-map layout associated with the comparison-map (left image).

\end{document}